\documentclass{article}

\usepackage[preprint,nonatbib]{neurips_2024_arxiv}

\usepackage[utf8]{inputenc} 
\usepackage[T1]{fontenc}    
\usepackage{hyperref}       
\usepackage{url}            
\usepackage{booktabs}       
\usepackage{amsfonts}       
\usepackage{nicefrac}       
\usepackage{microtype}      
\usepackage{xcolor}         

\usepackage{csquotes} 
\usepackage{etoolbox}
\usepackage{amsmath}
\usepackage{amssymb}
\usepackage[utf8]{inputenc}
\usepackage{wrapfig}
\usepackage{makecell}
\usepackage{multirow}
\usepackage{tcolorbox}
\usepackage{floatrow}
\newfloatcommand{capbtabbox}{table}[][\FBwidth]
\usepackage{listings}
\definecolor{deepblue}{rgb}{0,0,0.6}
\definecolor{deepred}{rgb}{0.6,0,0}
\definecolor{deepgreen}{rgb}{0,0.5,0}
\lstdefinestyle{python}{
    language=Python,
    basicstyle=\ttfamily\small,
    commentstyle=\color{deepred},
    otherkeywords={},             
    keywordstyle=\color{deepgreen},
    emph={},          
    emphstyle=\color{deepblue},    
    stringstyle=\color{deepred},
    showstringspaces=false            %
}
\usepackage{tablefootnote}
\usepackage{bbding}
\usepackage[capitalize]{cleveref}
\crefname{equation}{Eq.}{Eqs.}

\usepackage{graphicx}

\hypersetup{
    colorlinks=true,
    linkcolor=magenta,
}

\newtoggle{arxiv}

\renewcommand{\paragraph}[1]{\noindent\textbf{#1}}

\definecolor{darkgreen}{rgb}{0, 0.4, 0}
\definecolor{darkyellow}{rgb}{0, 0.4, 0.4}
\definecolor{brightred}{rgb}{1, 0, 0}

\newcommand{\mute}[1]{}

\newcommand{\iparagraph}[1]{\textbf{#1.} }

\newcommand{\kv}{$\akey{}\avalue{}$}
\newcommand{\kvcache}{\kv{}-cache}

\makeatletter
\DeclareRobustCommand\onedot{\futurelet\@let@token\@onedot}
\def\@onedot{\ifx\@let@token.\else.\null\fi\xspace}

\newcommand{\method}[1]{\textsc{#1}}

\newcommand{\ours}{\method{Live2Diff}}
\newcommand{\streamdiffusion}{\method{StreamDiffusion}}
\newcommand{\rerender}{\method{Rerender}}
\newcommand{\animatediff}{\method{AnimateDiff}}
\newcommand{\freenoise}{\method{FreeNoise}}
\newcommand{\flowvid}{\method{FlowVid}}
\newcommand{\streamingllm}{\method{StreamingLLM}}
\newcommand{\controlnet}{\method{ControlNet}}
\newcommand{\stablediffusion}{\method{StableDiffusion}}
\newcommand{\sdedit}{\method{SDEdit}}

\newcommand{\dreambooth}{\method{DreamBooth}}
\newcommand{\lora}{\method{LoRA}}

\newcommand{\tinyvae}{\method{Tiny-VAE}}
\newcommand{\lcmlora}{\method{LCM-LoRA}}

\newcommand{\reals}{\mathbb{R}} 

\newcommand{\latent}{z} 
\newcommand{\dtime}{t} 
\newcommand{\findex}{f} 
\NewDocumentCommand\finterval{O{1}O{2}m} {{\findex_{#1} : \findex_{#2}}} 
\newcommand{\uweights}{\theta} 
\newcommand{\enoise}{\epsilon} 
\newcommand{\texte}{\mathcal{T}} 
\newcommand{\txt}{c} 
\newcommand{\sample}{x} 
\newcommand{\encoder}{\mathcal{E}} 

\newcommand{\features}{f} 
\newcommand{\out}{\text{out}} 
\newcommand{\tin}{\text{in}} 
\DeclareMathOperator{\softmax}{softmax} 

\newcommand{\aquery}{Q} 
\newcommand{\akey}{K} 
\newcommand{\avalue}{V} 
\newcommand{\transpose}[1]{#1^\top} 
\newcommand{\numc}{C} 
\newcommand{\weights}{\mathcal{W}} 

\newcommand{\numwarmup}{L_w} 

\newcommand{\dencoder}{E^{cond}} 
\newcommand{\depthcond}{y} 
\newcommand{\numdsteps}{T} 
\newcommand{\numcontext}{L} 

\newcommand{\best}[1]{\textbf{#1}} 
\newcommand{\second}[1]{\underline{#1}} 
\newcommand{\up}{$\uparrow$} 
\newcommand{\down}{$\downarrow$} 

\newcommand{\frameorstep}{frame}

\newcommand\blfootnote[1]{
    \begingroup
    \renewcommand\thefootnote{}\footnote{#1}
    \addtocounter{footnote}{-1}
    \endgroup
}

\title{
Live2Diff: Live Stream Translation via Uni-directional Attention in Video Diffusion Models
} 

\author{%
  Zhening Xing\textsuperscript{\rm 1}\quad
  Gereon Fox\textsuperscript{\rm 2}\quad
  Yanhong Zeng\textsuperscript{\rm 1}\quad
  Xingang Pan\textsuperscript{\rm 3}\\
  \textbf{Mohamed Elgharib}\textsuperscript{\rm 2}\quad
  \textbf{Christian Theobalt}\textsuperscript{\rm 2}\quad
  \textbf{Kai Chen}\textsuperscript{\rm 1,$\dagger$}
  \\
  \vspace{-0.6em}\\
  $^1$Shanghai Artificial Intelligence Laboratory \\
  $^2$Saarland Informatics Campus, Max Planck Institute for Informatics\\
  $^3$S-Lab, Nanyang Technological University\\ \\
  \href{https://live2diff.github.io}{https://live2diff.github.io}\\
}

\begin{document}
\blfootnote{\textsuperscript{$\dagger$} denotes corresponding author.} 

\maketitle
\newcommand{\figoverview}{
    \begin{figure*}[t]
        \includegraphics[width=\linewidth]{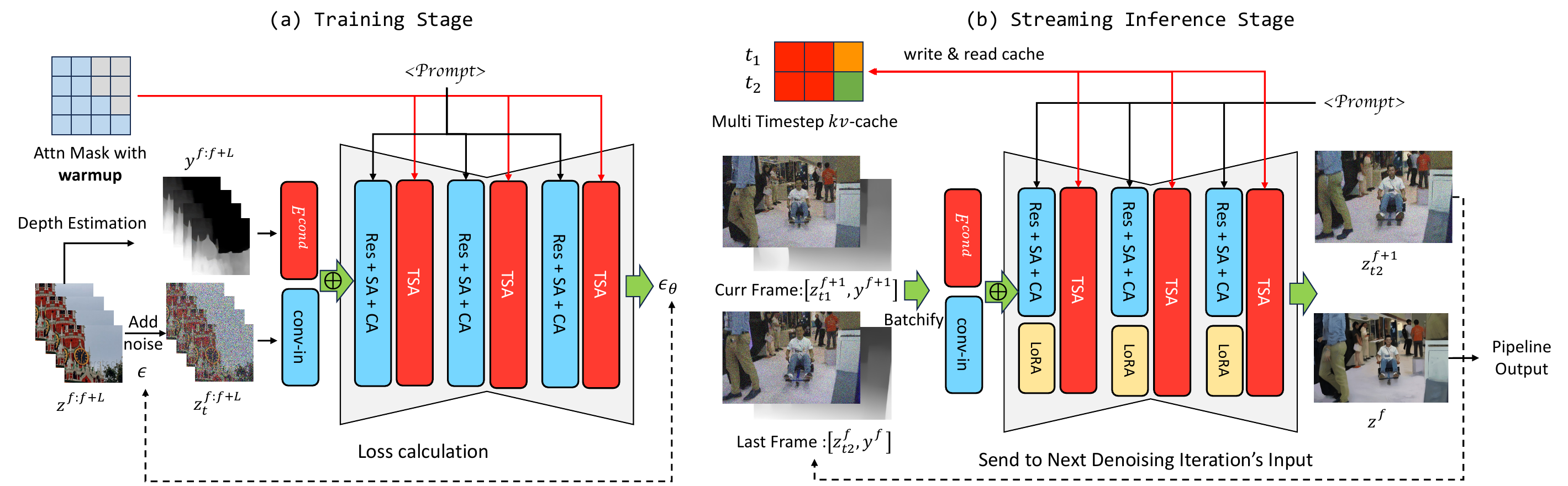}
        \caption{
            \textbf{The overview of \ours{}.}
            (a) During training, our model takes as inputs $L$ frames of noisy latents $\latent_\dtime^{f:f+L}$ and depth conditioning $\depthcond^{f:f+L}$, 
            where $f:f+L$ delimits the frame interval in a video stream, $\dtime$ is the denoising timestep, $\oplus$ denotes point-wise addition. 
            (b) During inference, frame $z^{f+1}$ is incorporated into the processing batch as it streams in, often before earlier frames are fully denoised. This results in a batch that includes frames at various denoising timesteps (e.g., $t_1$ and $t_2$). We employ a \kvcache{} mechanism to effectively reuse $\akey$ and $\avalue$ maps from previous frames, significantly improving inference efficiency while ensuring temporal consistency.
        }
        \label{fig:overview}
    \end{figure*}
}

\newcommand{\figattn}{
    \begin{figure}[t]
        \includegraphics[width=\linewidth]{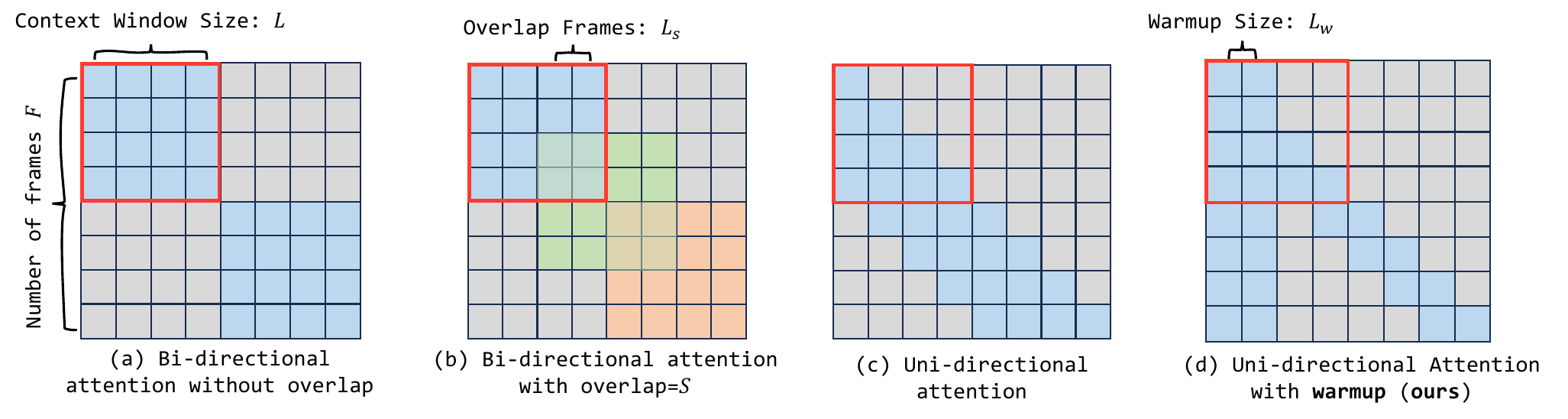}
        \caption{
                We visualize different types of temporal self-attention when the number of frames ($F = 8$) exceeds the length of the context window ($L = 4$). 
                The $j$-th cell of the $i$-th row is highlighted if the output for frame $i$ may contain information from frame $j$. The red square delineates the attention mask used during training.
                (a) shows temporal self-attention in current video diffusion models, which is bi-directional within the context window without overlap between chunks.
                (b) uses a sliding window with overlap $L_s$ (three subsequent positions of which are highlighted in different colors, for clarity) and fuses the output of overlap regions.
                (c) denotes the uni-directional attention widely used in LLMs.
                (d) shows the attention proposed by our method. We set the initial $L_w$ frames as warmup frames and apply bi-directional attention to them, while using uni-directional attention for the subsequent frames.
                The initial warmup frames also contribute to the output for all future frames.
        }
        \label{fig:attn_mask}
    \end{figure}
}

\newcommand{\figcompattn}{
    \begin{figure}[t]
        \begin{center}
            \includegraphics[width=0.95\textwidth]{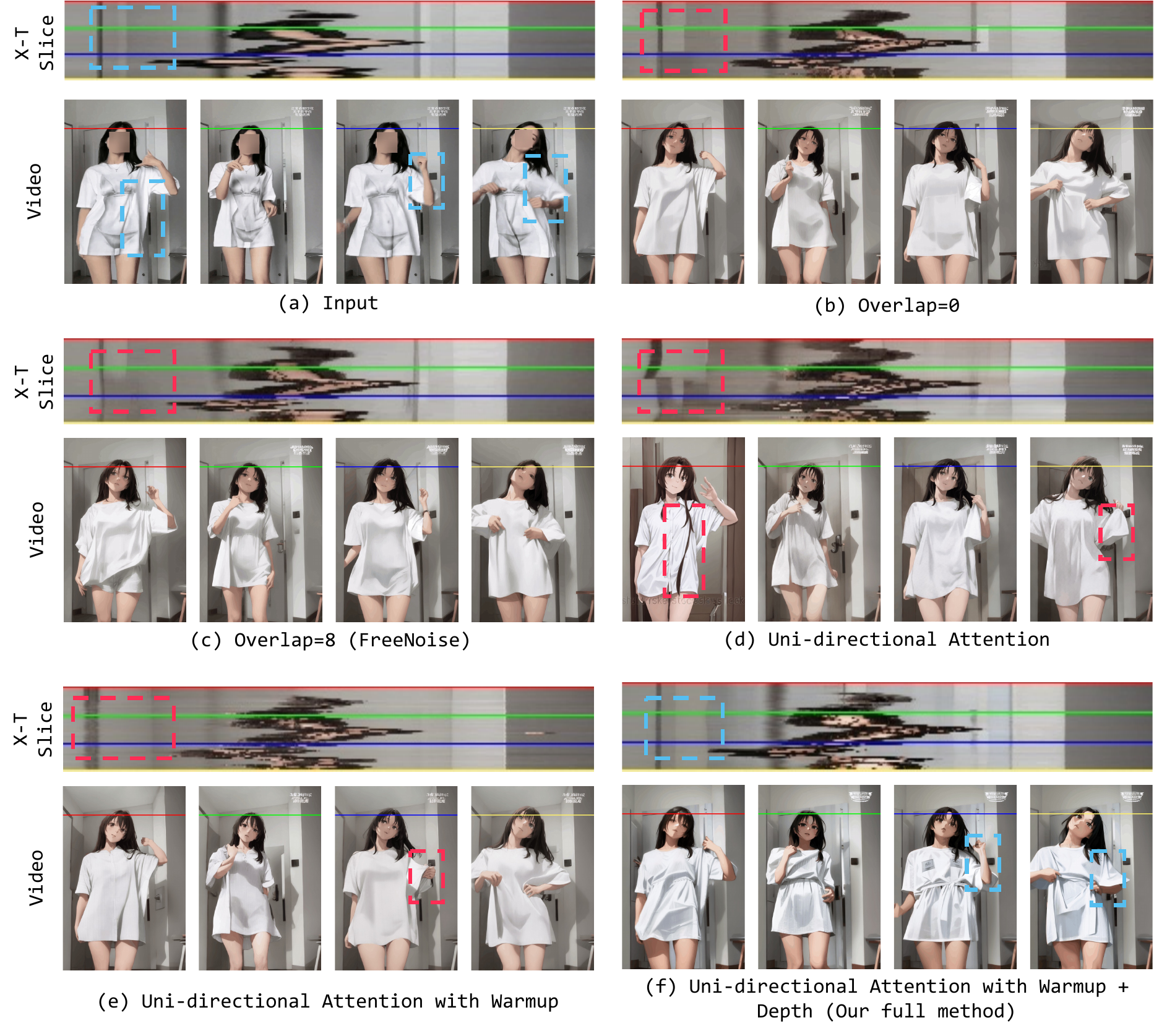} 
        \end{center}
        \caption{
            The X-T slice shows how the pixel values at the same X-coordinate change over time T. 
            The position of the horizontal lines in the video corresponds to the X-coordinate positions visualized in the X-T slice. 
            The color of each line represents the time in the X-T plot. 
            Red dashed boxes denote regions suffering from flickering and structural inconsistency, while blue boxes indicate areas where these issues are resolved.
            Flickering and gradual change in the background region can be observed in (b), (c) and (d), which use the first three attention modes illustrated in \cref{fig:attn_mask} respectively.
            In case (e), with the last attention mode from \cref{fig:attn_mask} (see also \cref{meth:tsa}, background flickering is reduced. 
            The depth conditioning in (f) improves structure consistency further.
        }
        \label{fig:comp_attn}
    \end{figure}
}

\newcommand{\figquality}{
    \begin{figure}
        \includegraphics[width=\linewidth]{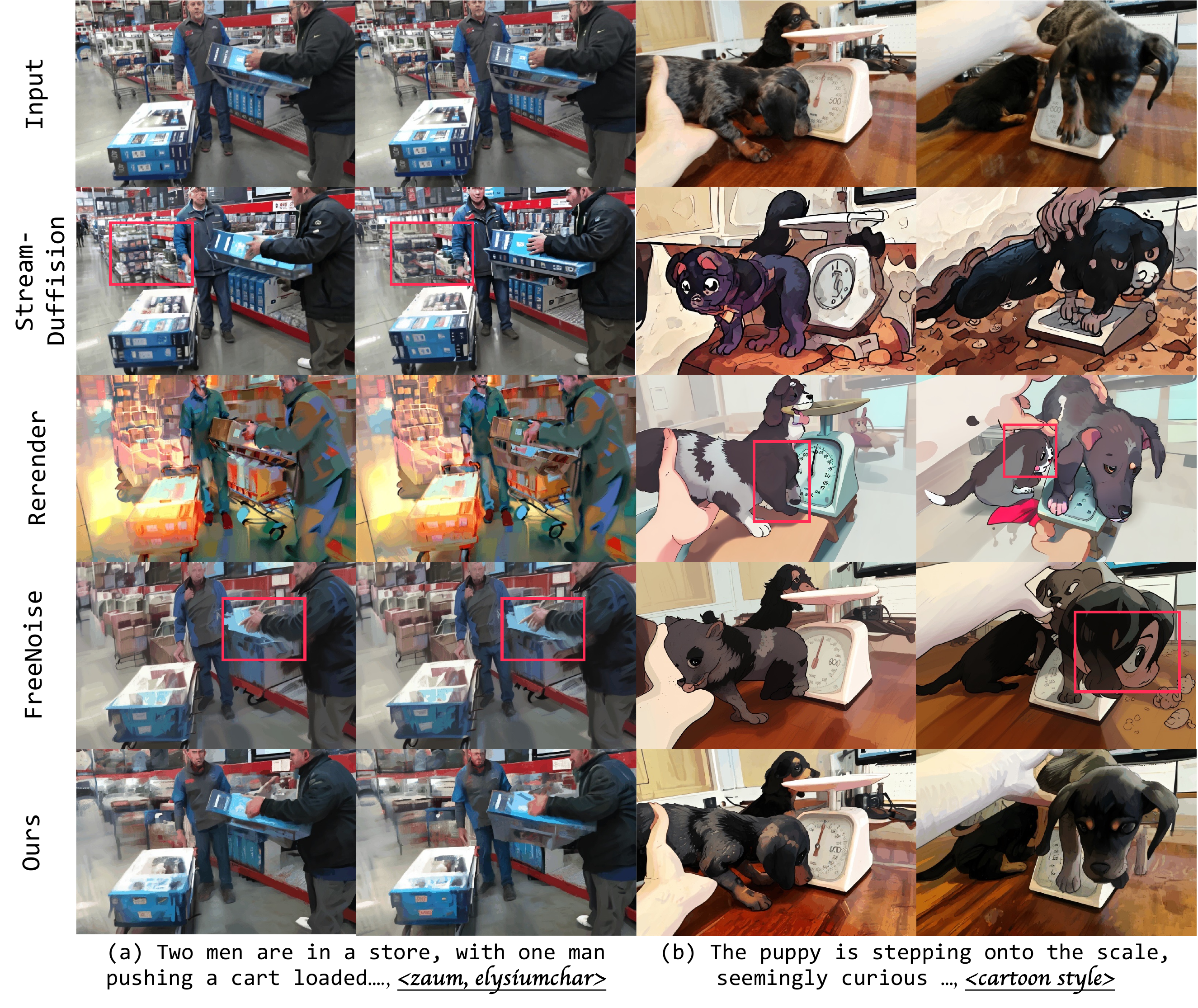}
        \vspace{-2mm}
        \caption{ We compare the output quality of our method to a number of previous approaches: (a) shows temporally adjacent frames, while (b) shows frames temporally further apart.  While our method preserves the spatial structure of the input well, producing the desired output styles, previous methods tend to change even the semantic content of the frames. See more discussions in \cref{sec:comparison}
        }\label{fig:qual}
        \vspace{-5mm}
    \end{figure}

}

\newcommand{\tabquantity}{
    \begin{table}[t]
        \begin{center}
        \setlength{\tabcolsep}{2pt}
        \begin{tabular}{lcccccc}
        \toprule
        \multirow{3}{*}{Method} & \multicolumn{2}{c}{Structure Consistency} & \multicolumn{3}{c}{Temporal Smoothness} & \multirow{2}{*}{Latency} \\ 
        \cmidrule(lr){2-6}
                        &                         Depth       &                            Ours         &                         CLIP      &            Warp                       &                             Ours          &           \\  
                        & \textcolor{white}{\down} MSE \down  & \textcolor{white}{\down} Win Rate \up &  \textcolor{white}{\up} Score \up & \textcolor{white}{\down} Error \down  &  \textcolor{white}{\down} Win Rate \up  &           \\  
                        \hline
        StreamDiffusion & \second{1.72}  & 70.4\%        & 94.35          & 0.1994          & 79.1\% &  \best{0.03s} ($\pm 0.01$)   \\    
        Rerender        & 5.50           & 86.7\%        & 95.05          & 0.1327          & 86.7\% &  7.72s ($\pm 0.18$)    \\  
        FreeNoise       & 1.94           & 80.5\%        & \best{96.49}   & \best{0.0809}   & 80.5\% &  58.67s ($\pm 0.08$)    \\ 
        Ours            & \best{1.12}    & -             & \second{95.77} & \second{0.0967} & -      &  \second{0.06s}  ($\pm 0.02$)    \\  
        \bottomrule
        \end{tabular}
        \end{center}
        \vspace{-3mm}
        \caption{
            To compare our method to previous work, we averaged scores over 90 sequences from the DAVIS-2017 \cite{pont2017davis} dataset.
            Our method scores \best{highest} in Depth MSE and \second{second} in terms of temporal smoothness. However, because \freenoise{} is actually unable to produce output frames before having seen a number of future input frames, we had to give it an unfair advantage by having it consume \emph{all} input frames before producing its first output frame, leading to extreme latency and explaining why it can achieve better temporal smoothness than all other methods. More details of the metrics in \cref{sec:evalsetup}. Our user study win rates confirm that our method produces the best quality for both aspects (i.e. all win rates over 50\%). Only \streamdiffusion{}, which puts more emphasis on speed than on output quality (see also \cref{fig:qual}) can beat our method in terms of latency.
        }
        \label{tab:quant}\label{tab:quantspeedmem}

    \end{table}
}

\newcommand{\tababl}{
    \begin{table}[t]
        \begin{center} 
            \setlength{\tabcolsep}{4pt}
            \begin{tabular}{ccccccc}
            \toprule
                \multicolumn{4}{c}{Setting} &  Structure Consistency & \multicolumn{2}{c}{Temporal Smoothness} \\
            
            \cmidrule(lr){1-4}
            & \makecell{Train \\ with warmup} & \makecell{Inference \\ with warmup} & \makecell{Use \\ depth} &    Depth MSE \down     &   CLIP Score \up & Warp error \down   \\
            \midrule
            A &     $\times$   & $\times$   & $\times$   & 2.29          & 95.43          & 0.0968          \\
            B &     \checkmark & $\times$   & $\times$   & 2.39          & 95.28          & 0.1125          \\
            C &     \checkmark & \checkmark & $\times$   & \second{2.22} & \best{95.80}   & \second{0.0966} \\
            D &     \checkmark & \checkmark & \checkmark & \best{1.67}   & \second{95.78} & \best{0.0768}   \\
            \bottomrule
            \end{tabular}
        \end{center}
        \caption{
            Ablation study of the model design. Our full method (D) achieves the \best{optimal} in both  structure consistency and temporal smoothness warp error.  As is to be expected, training with warmup, but filling the warmup area with immediate predecessor frames at test time (B) makes quality worse, but using the warmup area correctly (C) does lead to slight improvements over no warmup at all. The depth prior leads to a strong improvement again (D).     
        }
        \vspace{-4mm}
        \label{tab:abl}
    \end{table}
}

\newcommand{\figkvcache}{
    \begin{figure}
        \includegraphics[width=1\textwidth]{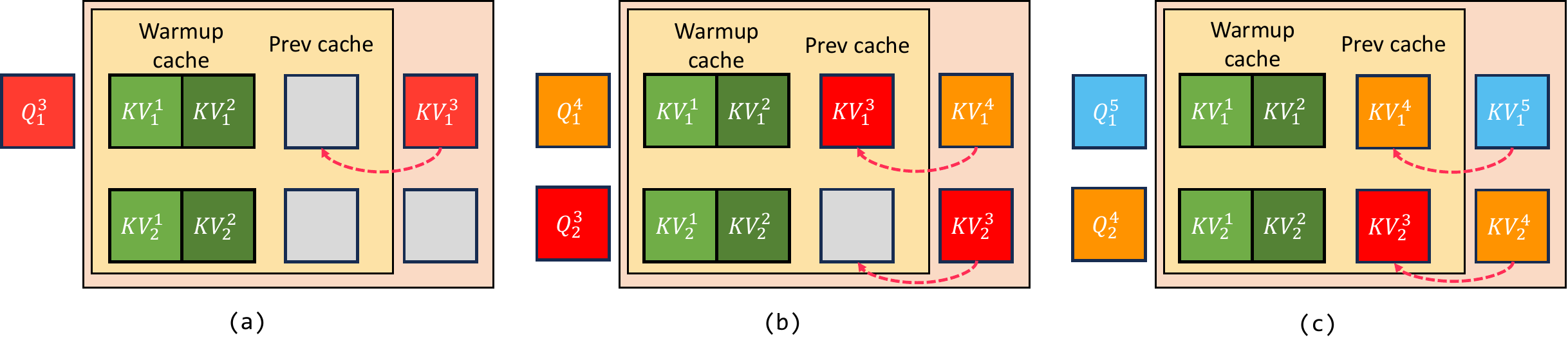}
        \caption{ (a) - (c) depict the usage of our \kvcache{} for the first steps of a stream, with $\numwarmup = \numdsteps = 2$. The colors of the squares indicate which frame they belong to. $\aquery, \akey, \avalue$ are the matrices used in \cref{equ:attn}, with subscripts indicating which denoising step they are used in and superscripts indicating which frame they belong to. Each row belongs to one of the two denoising steps. Red arrows are overwrite operations. For a step by step walkthrough see \cref{meth:pipeline}. 
        }
        \label{fig:kvcache}
    \end{figure}
}

\newcommand{\figabl}{
    \begin{figure}
    \vspace{-8mm}
        \includegraphics[width=\linewidth]{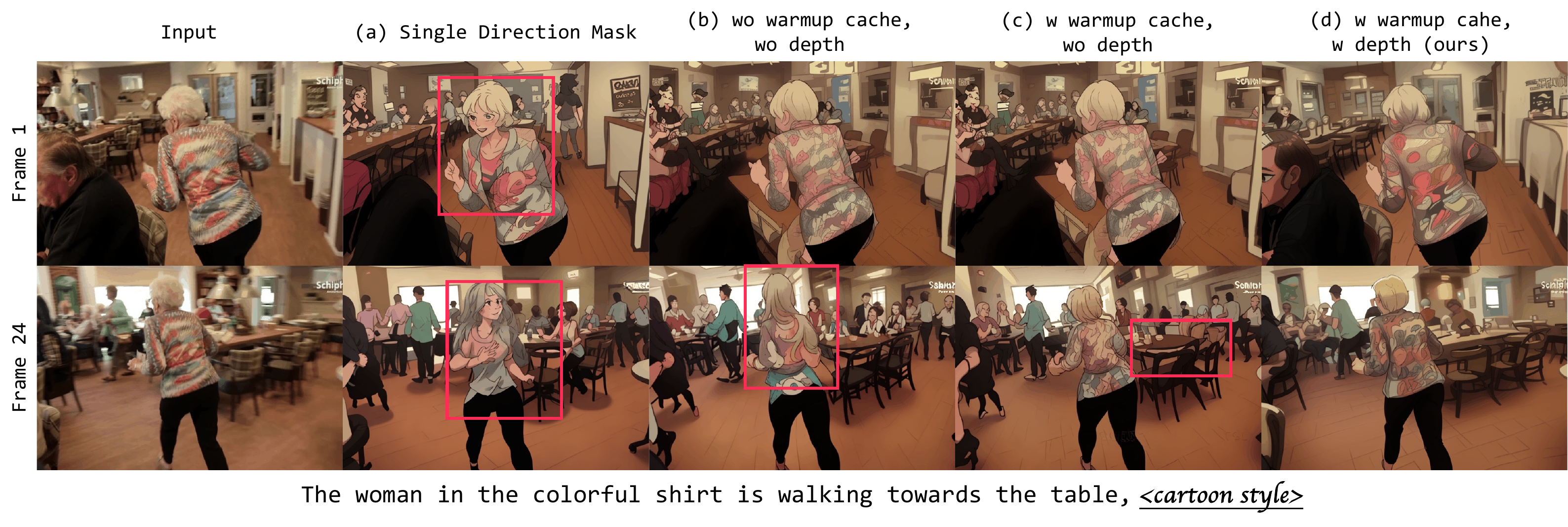} 
        \vspace{-6mm}
        \caption{In this ablation study, model (a) was trained with an attention mask like in \cref{fig:attn_mask} (c). Models (b) and (c) were trained with the attention masks like in \cref{fig:attn_mask} (d), but in (b) we filled the warmup slots (see \cref{fig:kvcache}) with further close-by predecessor frames instead of initial frames of the stream. Only (d), our full method, uses the depth prior.
        The models mostly agree on Frame 1, but all ablated versions deviate from the spatial structure of the input for later frames. More analysis in \cref{sec:ablation}.}
        \label{fig:abl}
        \vspace{-2mm}
    \end{figure}
}

\newcommand{\tablatency}{
    \begin{wraptable}[8]{r}{0.4\textwidth}
        \vspace{-0.35cm}
        \begin{tabular}{lc}
            \toprule
            Method        & Latency \down      \\ 
            \hline
            wo \kvcache{} & 20.43 ($\pm 0.068$) \\
            Ours          & 0.06s ($\pm 0.002$) \\ 
            \bottomrule
        \end{tabular}
        \caption{ Removing the \kvcache{} from our method drastically increases latency.
        }\label{tab:ablspeedmem}
    \end{wraptable}
}

\begin{abstract}

Large Language Models have shown remarkable efficacy in generating streaming data such as text and audio, thanks to their temporally uni-directional attention mechanism, which models correlations between the current token and \emph{previous} tokens.
However, video streaming remains much less explored, despite a growing need for live video processing.
State-of-the-art video diffusion models leverage
\emph{bi}-directional temporal attention to model the correlations between the current frame and all the \emph{surrounding} (i.e. including \emph{future}) frames, which hinders them from processing streaming videos.
To address this problem, we present \textbf{\ours{}}, the first attempt at designing a video diffusion model with uni-directional temporal attention, specifically targeting live streaming video translation.
Compared to previous works, our approach ensures temporal consistency and smoothness by correlating the current frame with its predecessors and a few initial warmup frames, without any future frames.
Additionally, we use a highly efficient denoising scheme featuring a \kvcache{} mechanism and pipelining, to facilitate streaming video translation at interactive framerates.
Extensive experiments demonstrate the effectiveness of the proposed attention mechanism and pipeline, outperforming previous methods in terms of temporal smoothness and/or efficiency.

\end{abstract}


\section{Introduction}
\label{sec:intro}
Large Language Models (LLMs) have recently been very successful in natural language processing and various other domains \cite{jiang2023mistral,jiang2024mixtral,touvron2023llama,touvron2023llama2,vaswani2017attention}.
At the core of LLMs is the autoregressive next-token prediction, seamlessly enabling real-time streaming data generation.
Such a mechanism processes data continuously as it streams in, bypassing the need for batch storage and delayed processing. 
This token prediction mode has been widely used in many applications such as dialog systems \cite{claude,openaiChat}, text-to-speech \cite{wang2023neural,zhang2023speak}, audio generation \cite{copet2024simple,huang2023make,kreuk2022audiogen}, etc.
Despite the recent success of this streamed generation of sequential data like text and audio, it has not been fully explored for another very common sequential data type: videos.
However, generating videos in a streaming manner is clearly worth investigating given the growing practical demand, particularly in live video processing, where the original input frames need to be translated into a target style on the fly.

Motivated by this, we study  next-frame-prediction for producing streaming videos, with streaming video-to-video translation as the target application.
Most existing video diffusion models exploit temporal self-attention modeling in a \emph{bi}-directional manner, 
where the models are trained on sequences of input frames to capture the pairwise correlations between all frames in a sequence \cite{blattmann2023stable,blattmann2023align,geyer2023tokenflow,guo2023animatediff,gupta2023photorealistic,yang2023rerender}. Despite some promising results, these models have limitations for streaming video. Early frames in a sequence rely on information from later frames, and vice versa for those at the end, which impedes efficient real-time processing of each frame as it streams in.

To address this issue, we redesign the attention mechanism of video diffusion models for streaming video translation, ensuring both high efficacy and temporal consistency. First, we make temporal self-attention \emph{uni}-directional via an attention mask, akin to attention in LLMs \cite{jiang2023mistral,vaswani2017attention}.
This removes the dependency of early frames on later frames in both training and inference,
making the model applicable to streaming videos.
However, ensuring both high inference efficiency and performance on long video streams using unidirectional attention is non-trivial. 
An approach is to use dense attention with all previous frames for next-frame prediction. However, it increases time complexity and decreases performance when the video length exceeds the attention window size used during training. 
Another option is to limit temporal self-attention to a smaller fixed window size during inference.
Unfortunately, unlike the sufficient context attention from user input tokens 
in LLM \cite{jiang2023mistral,vaswani2017attention,xiao2023streamingllm}, it is difficult to generate satisfactory frames with limited context attention at the beginning of the video stream, which further results in artifacts in later frames. 
To tackle this, we introduce warmup area in the unidirectional attention mask, which incorporates bi-directional self-attention modeling to compensate for the limited context attention at the beginning of the stream. During inference, we include the attention from a few warmup frames at the start of the stream to the current frame. Such a tailored attention design ensures both stream processing efficacy and temporal consistency modeling.

Building upon our tailored attention mechanism, we present \textbf{\ours{}}, a pipeline that processes \textbf{LIVE} video streams by a uni-directional video \textbf{DIFFUSION} model while ensuring high efficacy and temporal consistency.
First, our attention modeling mechanism removes the influence of later frames on previous frames, allowing for the reuse of $\akey$ and $\avalue$ maps from previously generated frames. This eliminates the need for recomputation when processing subsequent frames. We carefully designed a \kvcache{} feature in the diffusion pipeline to cache and reuse K/V maps, resulting in significant computation time savings.
Second, we further include a lightweight depth prior in the input, ensuring structural consistency with the conditioning stream.
Finally, \ours{} uses the batch denoising strategy to further improve stream processing efficacy, achieving 16FPS for $512\times 512$ videos on an RTX 4090 GPU.
We conduct extensive experiments to validate the superiority of \ours{} in terms of temporal smoothness and/or efficiency. We summarize our main contributions as follows,
\begin{itemize}
    \item To the best of our knowledge, we are the first to incorporate uni-directional temporal attention modeling into video diffusion models for video stream translation.  
    \item We introduce a new pipeline \textbf{\ours{}}, which aims at achieving live stream video translation with both high efficacy (16FPS on an RTX 4090 GPU) and temporal consistency. 
    \item We conduct extensive experiments including both quantitative and qualitative evaluation to verify the effectiveness of \ours{}. 
\end{itemize}

\figattn

\section{Related Work}
\label{sec:related}

\iparagraph{Attention}
LLMs~\cite{jiang2023mistral,jiang2024mixtral,touvron2023llama,touvron2023llama2} owe their success largely to the remarkable efficacy of the attention mechanism~\cite{vaswani2017attention}. 
In order to support the auto-regressive prediction of the next token, they use a uni-directional (or \enquote{masked}) attention mechanism, restricting the model to learning the dependence of \emph{later} tokens on \emph{earlier} ones, but no dependence of earlier tokens on later ones.
However, for tasks with very long token sequences, relating  the current token to all previous tokens becomes intractable.
To address this, \streamingllm{} \cite{xiao2023streamingllm} proposes to relate the current token to several initial tokens and a number of most recent tokens, which improves efficiency in handling long tokens.
While such kind of uni-directional attention is widely used in generating text and audio, video generation has not yet followed this trend: Bi-directional attention without masks is commonly used in video diffusion models~\cite{blattmann2023stable,blattmann2023align,guo2023animatediff,gupta2023photorealistic} to generate video chunks.
In this work, we study the use of uni-directional temporal attention in video diffusion models.
While our method draws inspiration from \streamingllm{}, it is the first time that such a design is studied in the video domain.

\iparagraph{Video Diffusion Models}
The multitude of possible conditioning modalities has made diffusion models the basis for image editing approaches \cite{meng2021sdedit,kawar2023imagic}, as well as video generation models \cite{guo2023animatediff,liang2023flowvid,kodaira2023streamdiffusion}. For example, \animatediff{} \cite{guo2023animatediff} extends \stablediffusion{} by a so-called \enquote{motion module}, enabling the denoising of entire video chunks based on temporal self-attention \cite{vaswani2017attention}. 
\freenoise{} \cite{qiu2023freenoise} is a method based on pretrained video diffusion models (e.g. \animatediff{} \cite{guo2023animatediff}) for long video generation.
This method carefully selects and schedules the latent noise for every time step in order to improve temporal smoothness. However, \freenoise{}, according to their experiments section produces frames at under 3FPS on an NVIDIA A100 GPU, which is not acceptable in the kinds of live streaming scenarios that we aim at (see \cref{sec:intro}). \flowvid{} \cite{liang2023flowvid} and \rerender{} \cite{yang2023rerender} produce frames at even lower rates, albeit with acceptably smooth results.

\iparagraph{Accelerating Diffusion Models}
Some recent diffusion-based methods \cite{luo2023lcm,luo2023lcmlora,song2023consistency,kodaira2023streamdiffusion} have prioritized low latency and/or high throughput: (Latent) consistency models (LCMs) \cite{song2023consistency,luo2023lcm} have reduced the number of denoising steps from 50 (the default in \stablediffusion{}) to as low as 4, leading to large speed ups without too much loss in quality. This principle has even been combined with the use of low-rank matrices for fine-tuning \cite{luo2023lcmlora}, allowing further speedup. A work that very specifically targets the streaming frame-by-frame translation setting is \streamdiffusion{} \cite{kodaira2023streamdiffusion}: Not only is this technique utilizing the aforementioned low-rank-adapted LCMs, but also it denoises video frames in a \enquote{pipelined} manner for the streaming scenario, i.e. the batch of images to be denoised can contain different levels of remaining noise, allowing new frames to be added to the batch before previous frames in the batch have been completely denoised, which makes optimal use of GPU parallelization.
However, \streamdiffusion{} renders videos frame-by-frame without any temporal modeling, leading to significant temporal discontinuity, which our method avoids due to the temporal correlations learned during training.

\section{Method}
\label{sec:method}

Our method, \ours{} (see \cref{fig:overview}), takes as input a stream of video frames, along with a matching text prompt. It produces a stream of output video frames, the spatial structure of which is similar to that of the input frames, but the appearance/style of which conforms to a specified target style, captured by \method{DreamBooth} \cite{ruiz2022dreambooth}. 
To achieve this, we replace the bidirectional temporal attention used in previous approaches by \emph{uni}directional attention (\cref{meth:tsa}). This allows us to cache $\akey$ and $\avalue$ maps from previous frames, leading to increased throughput (\cref{meth:pipeline}). Furthermore we accelerate generation by pipelined denoising, i.e. multiple time steps with different levels of residual noise are denoised in parallel. By employing \lcmlora{}\cite{luo2023lcmlora} we can drastically reduce the number of necessary denoising steps, which also helps meet framerate criteria.
We stabilize the spatial structure of frames with lightweight depth injection.

\figoverview

\subsection{Preliminaries}\label{meth:pre}
\iparagraph{Diffusion Models}
Diffusion models \cite{ho2020ddpm,dhariwal2021ADM} aim at undoing the so-called \enquote{forward process}, that iteratively adds Gaussian noise to the representation of a sample from a distribution. To achieve this, \method{StableDiffuion} \cite{rombach2022high} trains a \method{U-Net} \cite{ronneberger2015unet} to estimate the noise component of a noisy latent representation of any given image. By repeatedly estimating remaining noise and removing (some of) this noise from the latent code, a purely Gaussian noise vector can iteratively be denoised to obtain a clean sample as follows: Given a noisy latent code $\latent_\dtime$, the U-Net parametrized by weights $\uweights$ computes the estimated noise $\enoise_\uweights (\latent_\dtime , \dtime, \texte(\txt))$, where $\texte(c)$ is the CLIP encoding \cite{radford2021clip} of a conditioning text string $\txt$. The less noisy latent code $\latent_{\dtime - 1}$ can then be approximated as 
\begin{equation}
    \latent_{\dtime - 1} \approx \lambda \cdot \latent_\dtime + \mu \cdot \enoise_\uweights (\latent_\dtime , \dtime, \texte(\txt))
    \label{equ:diffusion}
\end{equation}
where $\lambda, \mu \in \reals$ are constants derived from the noise schedule of the forward process \cite{song2020ddim}.
The U-Net is trained by sampling images $\sample$ from the training distribution, mapping them to latent codes $\latent_0 = \encoder(\sample)$ and then adding varying amounts of Gaussian noise to obtain $\latent_\dtime$, such that the U-Net output can be supervised by $L1$ distance to the known noise. Like in \method{StableDiffusion}, we use this as our main loss, but with $\sample$ holding not single images, but chunks of consecutive video frames.

\iparagraph{Bidirectional Attention in Video Generation}
Several video diffusion models \cite{guo2023animatediff,blattmann2023align,wang2023modelscope,chen2024videocrafter2} use temporal self-attention layers to improve temporal smoothness of the output, essentially encouraging the model to learn temporal correlations. 
A temporal self-attention layer computes its output  as
\begin{equation}\label{equ:attn}
    \features_\out := \softmax\left(\frac{\aquery \transpose{\akey}}{\sqrt{\numc}}\right) \cdot \avalue
\end{equation}
where $\aquery:=\weights_\aquery \cdot \features_\tin$, $\akey:=\weights_\akey \cdot \features_\tin$, and $\avalue:=\weights_\avalue \cdot  \features_\tin$ are linear projections of the input features$\features_\tin$ and $\numc$ is the number of feature channels. 
\method{AnimateDiff} adds sinusoidal position encoding to $\features_\tin$ before computing \cref{equ:attn}, to give the layer access to the temporal position of each feature vector.
Note that $\features_\tin$ contains information about an entire chunk of frames, as does $\features_\out$.
Since previous works use such temporal attention layers without masking \cite{vaswani2017attention}, $\features_\out$ can thus base its information about a particular frame on frames before and \emph{after} that frame. Exploiting temporal correlations in this bidirectional way helps produce temporally smooth output, but is counter-productive for the streaming setting, as a prefix of $\features_\out$ will often need to be computed before the full $\features_\tin$ is even available.
In \method{FreeNoise} \cite{qiu2023freenoise} this \enquote{direct} bidirectionality was constrained in the sense that only frames from a \enquote{temporal neighborhood} can influence each other \emph{directly} (see \cref{fig:attn_mask} (b)), but since the U-Net contains multiple temporal attention layers and because denoising requires multiple U-Net applications, the distance over which information can be transferred backwards in time is still considerable. \cref{fig:comp_attn} (b) and (c) show the artifacts of using this attention mode for streaming.

\figcompattn

\subsection{Input \& Structural prior}\label{meth:cond}

Following \animatediff \cite{guo2023animatediff}, we insert temporal self-attention modules (TSA) in-between the \stablediffusion \cite{rombach2022high} U-Net layers, see \cref{fig:overview}.
In order to condition our generated video on the input video similarly to \method{SDEdit} \cite{meng2021sdedit} we add a certain amount of Gaussian noise to the input frames. 
The strength of this noise is chosen such that the high level spatial structure of the image is still recognizable, but the appearance information is largely erased at finer scales. This requires the denoising U-Net to fill in appearance information according to the target style.
To facilitate the preservation of spatial structure we use an additional depth input: We use \method{MiDas} \cite{ranftl2022midas,ranftl2021midas} for frame-wise depth estimation. The depth frames are then encoded by \method{StableDiffusion}'s encoder $\encoder$, with the results being fed into a lightweight convolutional module $\dencoder$ (see \cref{fig:overview}).
Finally, we add the output of the conditional module to the first convolution layer and pass it through the U-Net.
We found  this explicit structural prior to help the transferred video maintain better structural consistency with the source video.
Evidence can be found in \cref{fig:comp_attn} and \cref{sec:results}.

\subsection{Uni-directional Temporal Self-Attention with Warmup}\label{meth:tsa}
To turn bi-directional attention, as shown in \cref{fig:attn_mask} (a), in which each \frameorstep{} inside a chunk can be based on information from all other time steps in the chunk, into uni-directional attention, where each \frameorstep{} can only depend on \emph{earlier} \frameorstep{}s, we use masked attention \cite{vaswani2017attention,jiang2023mistral}. 
\cref{fig:attn_mask} (c) presents a solution using the uni-directional attention mask commonly used in LLM\cite{touvron2023llama,touvron2023llama2, vaswani2017attention, jiang2024mixtral}.
However, as shown in red dashed box in \cref{fig:comp_attn} (d), the spatial structure of the first frame's output is inconsistent with the input, and the character identity differs from subsequent frames. Flickering in the background region can be observed in the X-T slice as well. We believe that the reason for this attention mode being less effective for video transfer than for LLMs is rooted in the fact that in LLMs, attention operations rely on user prompts as initial tokens for uni-directional attention, while in video transfer such initial tokens first need to be generated by the model.

This is why we choose the attention mask in \cref{fig:attn_mask} (d): We apply bi-directional attention in the corner case of the first $L_w$ \enquote{warmup} frames, and unidirectional attention afterwards.
In \cref{fig:comp_attn} (e) we observe that the flickering issue in the red dashed region has been resolved. 
The remaining structure inconsistencies between the red and blue dashed regions compared to the input are resolved by the structural prior (see \cref{meth:cond}).

\figkvcache

\subsection{High Efficiency Inference Pipeline}\label{meth:pipeline}
\iparagraph{\kvcache{}} As described in \cref{meth:tsa} our warmup-based temporal self-attention makes sure that when we compute the attention for a certain frame, the attention for all previous frames has already been computed. This means that those parts of the matrices $\akey$ and $\avalue$ in \cref{equ:attn} that concern the previous frames do not need to be computed again, but can be retrieved from a cache, our \kvcache{}. Note that the diffusion U-Net has multiple temporal attention layers, and that the U-Net needs to be applied $\numdsteps$ times in order to fully denoise a frame. This means that for every combination of layer and denoising step, we keep a separate \kvcache{}.
More details  in \cref{supp:kv_cache}.

\iparagraph{Warmup stage}
At the start of the stream, we feed $\numwarmup$ frames into the U-Net and denoise them completely, with bidirectional attention as introduced in \cref{meth:tsa}. This gives us the $\akey$ and $\avalue$ matrices highlighted in green in \cref{fig:kvcache}. They are used for \emph{all} future frames for temporal consistency.

\iparagraph{Pipelined denoising} Similarly to \method{StreamDiffusion} \cite{kodaira2023streamdiffusion}, we denoise frames in a pipelined manner, i.e. as soon as the next input frame becomes available, we add it to the batch of frames to be denoised, even though it may contain a much higher amount of noise than previous frames in the batch that have already undergone multiple denoising steps. This way we utilize our GPU capacity most efficiently, increasing throughput.

\cref{fig:kvcache} illustrates the bookkeeping in our \kvcache{} for the simple case $\numwarmup = \numdsteps = 2$, just after warmup: In the first step, to process the red feature vector, temporal self attention according to \cref{equ:attn} is computed only using the green $\akey$ and $\avalue$ maps and the red one. The $\akey$ and $\avalue$ for the input token are then written into the cache (red arrow) and one denoising step can be computed. In the second step, the orange frame arrives, which in addition to the green warmup tokens we also correlate with the previously cached $\akey$ and $\avalue$ for the red frame. In parallel with the first denoising step for the orange frame, we compute the second denoising step for the red frame. Now all caches are filled, for the third step, in which the blue input frame arrives.

\subsection{Training}\label{meth:training}
\vspace{-2mm}
To train our model we use data collected from Shutterstock \cite{shutterstock2024}, resized to resolution $256 \times 256$. We choose $\numcontext = 16$ and $\numwarmup = 8$.
We train our model as follows: We initialize the weights of our temporal self-attention modules with the weights from \method{AnimateDiff} and fine-tune them for 3000 iterations using our uni-directional attention (\cref{meth:tsa}). Then we add $\dencoder$ (\cref{meth:cond}), with the last layer initialized with zeros \cite{zhang2023controlnet} and train all weights jointly for 6000 iterations.
We use the Adam \cite{kingma2014adam} optimizer with a learning rate of $1e-4$ and train on batches of 4 samples per GPU, on 8 GPUs. Accumulation of 32 gradients leads to an effective batch size of 1024.

\section{Results}
\label{sec:results}

\subsection{Evaluation Setup} \label{sec:evalsetup}
\vspace{-2mm}
\iparagraph{Dataset} 
We evaluate on the DAVIS-2017 \cite{pont2017davis} dataset, which contains 90 object-centric videos.
We resize all frames to $512\times 768$ via bilinear interpolation and use \method{CogVLM} \cite{wang2023cogvlm} to caption the middle frame of each video clip. 
To specify the target style, we add the corresponding trigger words of DreamBooth and LoRA as suffix. For more information about the prompts, see \cref{supp:eval}.

\iparagraph{Metrics}
We evaluate three aspects of the generated videos: \emph{structure consistency} (Output frames should have similar spatial structure as input frames), \emph{temporal smoothness} (no sudden jumps in the motion) and \emph{inference latency}. 
We measure \emph{structure consistency} as the mean squared difference between the depth maps estimated \cite{ranftl2021midas} for the input and output frames.
As in previous work \cite{wu2023tune,khachatryan2023text2video,guo2023animatediff} we measure \emph{temporal smoothness} by  CLIP score \cite{radford2021clip}, i.e. by the cosine similarity of the CLIP embeddings of pairs of adjacent frames. In addition we compute the so-called \enquote{warp error} \cite{lai2018warpError} for pairs of adjacent frames, i.e. we compute the optical flow \cite{teed2020} between the frames and then warp the predecessor frame accordingly, to compute a weighted MSE between the warping result and the successor.
We also conduct a user study to evaluate \emph{structure consistency} and \emph{temporal smoothness}: Each participant is given triplets of videos (original input video, result from our method, result from a random different method) and is asked to identify the result with the best quality. Then we calculate the rate of our method winning compared to other methods. A higher win-rate indicates that users perceive our method to be better in the corresponding evaluation direction. Please refers to \cref{supp:eval} for detailed information.
We measure \emph{inference speed} as the total amount of time it takes each method to process an input stream of 100 frames at resolution $512 \times 512$, on a consumer GPU (NVIDIA RTX 4090).

\subsection{Comparisons}\label{sec:comparison}
\vspace{-2mm}

We compare our method to three previous works, all based on \stablediffusion{} \cite{rombach2022high} and compatible with  \dreambooth{} \cite{ruiz2022dreambooth} and \lora{} \cite{hu2021lora}:
\textbf{\streamdiffusion}\cite{kodaira2023streamdiffusion} applies \sdedit{} on a frame-by-frame basis. The same noise vector is used for all the frames, to improve consistency and smoothness. To achieve interactive framerates, \streamdiffusion{} uses \lcmlora{}\cite{luo2023lcmlora} and \tinyvae{}\cite{ollin2022taesd}, with the latent consistency model scheduling \cite{luo2023lcm}. We apply the same acceleration techniques in our method.
\textbf{\rerender{}}\cite{yang2023rerender} first inverts input key frames into noisy latent codes. During denoising, temporal coherence and spatial structure stabilization are achieved by using cross-frame attention, flow-based warping and \controlnet{} \cite{zhang2023controlnet}. We select all frames as key frames for the purposes of our evaluation, but otherwise use the default settings.
\textbf{\freenoise{}} \cite{qiu2023freenoise} does not natively support an input video as conditioning, but by adding a sufficient amount of noise to the input, similar to our method and \sdedit{}, we can nevertheless use it for our video-to-video translation task. 
The amount of noise we add is equivalent to half of the entire denoising process.
\freenoise{} uses a technique called  window-based attention fusion, that (similar to bidirectional temporal attention) leads to frames incorporating information from future frames. This actually makes it unsuitable for the streaming setting, which we mitigate by giving \freenoise{} access to \emph{all} frames, not expecting to receive the first output frames after we have given the last input frames. In this sense we are giving \freenoise{} a considerable advantage.
\iparagraph{Qualitative Comparison}
\cref{fig:qual} compares outputs of all methods: While part (a) shows two consecutive output frames, part (b) shows frames that are further apart.
\streamdiffusion{}\cite{kodaira2023streamdiffusion} exhibits strong flickering in the background (red box in (a)). When foreground and background are difficult to distinguish (box and shelf in (a), dog and table in (b)), works other than ours struggle to produce satisfactory results: 
\streamdiffusion{} generates inconsistent results with low quality.
\rerender{} generates strong artifacts in the first frame (see (a)) and propagates them to later frames. 
\freenoise{} fails to adhere to the input frame and generates elements unrelated to prompt and input.
In contrast, our method leverages depth information to ensure the structural accuracy of the generated results (e.g. the box in \cref{fig:qual} (a)) and maintains consistency over longer duration (dog in \cref{fig:qual} (b)).

\iparagraph{Quantitative Comparison}
In \cref{tab:quant} we evaluate structure consistency and temporal smoothness:
While our method outperforms the others in structure consistency, we observe that \freenoise{} achieves a better CLIP score and warp error for temporal smoothness. This is not surprising, as the way that time steps are correlated in \freenoise{} allows information to flow bidirectionally along the temporal axis, which, unlike for all other methods, required \freenoise{} to be given access to \emph{all} frames at once (see first paragraph of \cref{sec:comparison}). This is an unfair advantage to \freenoise{}, violating some assumptions of the streaming scenario, as it allows \freenoise{} to correlate its output frames to input frames that would likely not be available at the time the output frame needs to be produced.
In our user study, all our win rates are way above 50\% for  both structural consistency and temporal smoothness, confirming that our results are the most convincing.
\streamdiffusion{}\cite{kodaira2023streamdiffusion} scores second-best in structure consistency, likely because it applies only a moderate amount of noise to its input, but this limits its ability to conform with the target style (see also \cref{fig:qual}).
\cref{tab:quantspeedmem} also compares inference latencies, i.e. the average time that goes by between receiving an input frame and producing the corresponding output frame. As is to be expected for a method that first consumes all the frames before producing any output, \freenoise{} has by far the largest latency, which makes it unusable for the live streaming scenario. Only \streamdiffusion{} has a better latency than our method, which we attribute to it making a different tradeoff between performance and quality. This is confirmed by the MSE , CLIP scores, warp error, as well as the user study results, that consistently indicate higher output quality for our method.

\tabquantity

\figquality

\figabl

\subsection{Ablation Study}\label{sec:ablation}

We include quality and quantity results of model with different setting in \cref{fig:abl} and \cref{tab:abl}.
We employ a noise strength of 0.5 for more apparent comparison.
The model (a), which uses uni-directional attention (see \cref{fig:attn_mask} (c)) fails to be consistent with the input from the first frame. 
Columns (b), (c) and (d) are trained with our uni-directional attention with warmup (red square of \cref{fig:attn_mask} (d)). But the model in (b) fills the warmup area with further predecessor frames instead of initial frames (see also \cref{fig:kvcache}). 
This does improve the output for the first frame (as the predecessor frames happen to be initial), but leads to deviation from the spatial structure of the input in later frames. 
\tababl
In column (c) we do use the warmup area properly, but omit the depth prior. 
The identity of the subject is now maintained better, but several details in the background, such as the highlighted table region are still inconsistent with the input.
Only our full method (column (d)) maintains consistency beyond initial frames.
\cref{tab:abl} confirms these findings: Without the depth prior, configurations A, B and C fail to be structurally consistent with the input. 
And with further predecessor frames instead of initial frames in the warmup area at inference time, configuration B does not achieve as much temporal consistency as the others. We also found that removing the warmup cache from configuration D will decrease the temporal smoothness CLIP score by $0.09$.
The depth prior seems to improve both structure consistency and temporal smoothness a lot, although the temporal CLIP score fails to show that in \cref{tab:abl}. We interpret this failure as a consequence of the content of subsequent frames being usually very similar, such that the CLIP embeddings can be similar despite certain abrupt changes, for example in the background, being present.

As reported in \cref{tab:ablspeedmem}, omitting our \kvcache{} leads to our method having to re-compute the $\akey$ and $\avalue$ maps of previous frames multiple times, which dramatically increases the per-frame latency to a degree that is not acceptable in streaming use cases. 

\tablatency

\vspace{-2mm}
\section{Conclusion}
\label{sec:conclusion}
\vspace{-3mm}
We have presented \ours{}, a method to translate video streams to a desired target style at interactive framerates. Based on our novel unidirectional attention approach, that allows us to reduce computational cost by means of our \kvcache{}, we are able to not only meet the criterion of sufficient framerate, but also outperform previous approaches in terms of consistency with the input video and temporal smoothness. We have thus demonstrated that the unidirectional temporal attention mode, that is an important component of state of the art LLMs, can beneficially be used for the editing of videos as well.
A method like ours  could be of great use in a number of video streaming use cases, such as in the recent trend of \enquote{Virtual YouTubers}, in which online content producers control stylized virtual avatars and interact with their audience in a live stream.

{
    \small
    \bibliographystyle{plain}
    \bibliography{references}
}

\appendix
\section{Appendix / Supplemental material}

\subsection{Limitations \& Societal impact}
\label{sec:limitations}
Although our method improves upon the state of the art, we sometimes observe flickering in background regions, especially when the camera is moving fast, which we hypothesize to be the result of imperfect depth estimation. Also, editing the video by applying Gaussian noise and then denoising it limits our method to translation tasks in which the output is structurally similar to the input, ruling out, for example, Pose-to-Person conditioning.

Like other editing works our method could in principle be misused to manipulate video data, which weakens the value of such data as a means of verifying that an event in question actually took place. Because of said limitation to structure-preserving transfers this danger is somewhat smaller than for other methods, but given the right kind of target style it could still be used for example to alter the identity of depicted persons.

\subsection{Implementation details of our inference pipeline}\label{supp:kv_cache}

In this section, we provide the implementation of our temporal self-attention module with \kvcache{}. We also describe how we apply streaming inference.

\iparagraph{\kvcache{}}
In model initialization, we pre-compute the shape of the \kvcache{} for each temporal self-attention module.
For temporal attention with max window size $L$, and input feature size $H\times W\times C$, for $T$ denoising steps, the shape of the \kvcache{} should be $(T, H\times W, L, C)$, see \cref{lst:cachesize}.

\begin{lstlisting}[style=python,caption={\kvcache{} sizing},label=lst:cachesize]
def set_cache(T, H, W, L, C):
    k_cache = zeros(T, H * W, L, C)
    v_cache = zeros(T, H * W, L, C)
    register_buffer("k_cache", k_cache)
    register_buffer("v_cache", v_cache)
\end{lstlisting}

Previous video diffusion models\cite{guo2023animatediff,chen2024videocrafter2,wang2023modelscope} apply absolute positional encoding $\mathtt{PE}$, which is added to the input features before the mapping layers $\mathtt{to\_q}, \mathtt{to\_k}, \mathtt{to\_v}$, which can be formulated as
\begin{align*}
    \mathtt{\aquery} & = \mathtt{to\_q} (\mathtt{PE} + \mathtt{feat}) \\
    \mathtt{\akey} & = \mathtt{to\_k} (\mathtt{PE} + \mathtt{feat}) \\
    \mathtt{\avalue} & = \mathtt{to\_v} (\mathtt{PE} + \mathtt{feat})
\end{align*}
We thus cannot directly cache $\mathtt{\akey{}, \avalue{}}$ since they contain  positional information. 
Instead, we pre-compute $\mathtt{to\_q}(\mathtt{PE}), \mathtt{to\_k}(\mathtt{PE}), \mathtt{to\_v}(\mathtt{PE})$ (see \cref{lst:pecomp}), and cache only $\mathtt{to\_k}(\mathtt{feat}), \mathtt{to\_v}(\mathtt{feat})$.

\begin{lstlisting}[style=python,label=lst:pecomp,caption={We precompute $\mathtt{to\_k}(\mathtt{PE}), \mathtt{to\_v}(\mathtt{PE})$.}]
def prepare_pe_buffer():
    pe_full = pos_encoder.pe
    q_pe = F.linear(pe_full, to_q.weight)
    k_pe = F.linear(pe_full, to_k.weight)
    v_pe = F.linear(pe_full, to_v.weight)

    register_buffer("q_pe", q_pe)
    register_buffer("k_pe", k_pe)
    register_buffer("v_pe", v_pe)
\end{lstlisting}


In the warmup stage, we use bi-directional attention over all warmup frames, and cache their $\mathtt{K/V}$, see \cref{lst:warmup}.

\begin{lstlisting}[style=python,label=lst:warmup,caption={Warmup frames are processed with \emph{bi}-directional attention.}]
def temporal_self_attn_warmup(feat, timestep):
    """
    feat: [HW, L, C_in]
    """
    q = to_q(feat)  #  [HW, L, C]
    k = to_k(feat)  #  [HW, L, C]
    v = to_v(feat)  #  [HW, L, C]

    # cache warmup frames before positional encoding
    k_cache[timestep, :, :warmup_size] = k
    v_cache[timestep, :, :warmup_size] = v

    pe_idx = list(range(k.shape[1]))
    
    pe_q = q_pe[:, pe_idx]
    pe_k = k_pe[:, pe_idx]
    pe_v = v_pe[:, pe_idx]

    q_full = q + pe_q
    k_full = k + pe_k
    v_full = v + pe_v

    # do not use attention mask
    feat = scaled_dot_product_attention(
        q_full,
        key_full,
        value_full,
        attention_mask=None)

    feat = to_out(feat)
    return feat
\end{lstlisting}

During streaming inference we process up to $T$ samples with different noise levels at once.
For each frame we write to and read from the \kvcache{} corresponding to its noise level, and add the mapped positional information.
At the beginning of the stream, the length of context window is incrementally approaching the max window size $L$. We pass an attention mask to specify which token should take part in attention. 
For details see \cref{lst:attention,lst:streaming}.

\begin{lstlisting}[style=python,label=lst:attention,caption={Our implementation of streaming inference uses the uni-directional attention approach, see \cref{fig:attn_mask} (d).}]
def temporal_self_attn_streaming(feat, attn_mask):
    """
    feat: [THW, L, C_in]
    attn_mask: [T, L], 0 for attention, -inf for no attention
    """
    q_layer = rearrange(q_layer, "(nhw) f c -> n hw f c", n=T)
    k_layer = rearrange(k_layer, "(nhw) f c -> n hw f c", n=T)
    v_layer = rearrange(v_layer, "(nhw) f c -> n hw f c", n=T)

    # handle prev frames, roll back
    k_cache[:, :, warmup_size:] = k_cache[:, :, warmup_size:] \
                                  .roll(shifts=-1, dims=2)
    v_cache[:, :, warmup_size:] = v_cache[:, :, warmup_size:] \
                                  .roll(shifts=-1, dims=2)
    # write curr frame
    k_cache[:, :, -1:] = k_layer
    v_cache[:, :, -1:] = v_layer

    k_full = k_cache
    v_full = v_cache

    # attn_mask:
    #   [[0, 0, 0, 0, -inf, -inf, 0, 0],
    #    [0, 0, 0, 0, -inf, -inf, -inf, 0]]
    # then pe for each element shoule be
    #   [[0, 1, 2, 3, 3, 3, 4, 5],
    #    [0, 1, 2, 3, 3, 3, 3, 4]]
    kv_idx = (attn_mask == 0).cumsum(dim=1) - 1  # [T, L]
    q_idx = kv_idx[:, -q_layer.shape[2]:]  # [T, 1]

    # [n, window_size, c]
    pe_k = concatenate([
        k_pe.index_select(1, kv_idx[idx]) 
        for idx in range(T)], dim=0) 
    pe_v = concatenate([
        v_pe.index_select(1, kv_idx[idx]) 
        for idx in range(T)], dim=0) 
    pe_q = concatenate([
        q_pe.index_select(1, q_idx[idx]) 
        for idx in range(T)], dim=0) 

    q_layer = q_layer + pe_q.unsqueeze(1)
    k_full = k_full + pe_k.unsqueeze(1)
    v_full = v_full + pe_v.unsqueeze(1)

    q_layer = rearrange(q_layer, "n hw f c -> (n hw) f c")
    k_full = rearrange(k_full, "n hw f c -> (n hw) f c")
    v_full = rearrange(v_full, "n hw f c -> (n hw) f c")

    attn_mask_ = attn_mask[:, None, None, :].repeat(
        1, h * w, q_layer.shape[1], 1)
    attn_mask_ = rearrange(attn_mask_, "n hw Q KV -> (n hw) Q KV")
    attn_mask_ = attn_mask_.repeat_interleave(heads, dim=0)
    
    feat = scaled_dot_product_attention(
        q_full,
        key_full,
        value_full,
        attention_mask=attention_mask_)

    feat = to_out(feat)
    return feat
    
\end{lstlisting}

\begin{lstlisting}[style=python,label=lst:streaming,caption={During inference we process $T$ frames simultaneously to make full use of GPU parallelization.}]
def streaming_v2v(frame):
    """
    frame: [1, 3, H, W]
    """
    latent = vae.encode(frame)
    depth_latent = vae.encode(depth_detector(frame))
    noisy_latent = add_noise(latent)  # add noise based on SDEdit
    if prev_latent is None:
        prev_latent = randn([T-1, ch, h, w])
    if attn_mask is None:
        attn_mask = zeros(T, L)
        attn_mask[:, :warmup_size] = 1
        attn_mask[0, -1] = 1  # curr frame participate attention
        attn_mask.masked_fill_(attn_mask == 0, float("-inf"))
    
    latent_batch = concatenate([noisy_latent, prev_latent])
    noise_pred = UNet(latent_batch, depth_latent, 
                      t, text_embedding, attn_mask)
    denoised_latent = scheduler.step(noise_pred, latent_batch, t)
    
    out_latent = denoised_latent[-1]
    prev_latent = denoised_latent[1:]

    out_frame = vae.decode(out_latent)
    return out_frame
\end{lstlisting}

\begin{table}[h]
    \centering
    \begin{tabular}{cc}
        \toprule
         Models & Trigger Words  \\
         \hline
         Flat-2D Animerge \tablefootnote{https://civitai.com/models/35960/flat\-2d\-animerge} & \textit{cartoon style} \\
         zaum \tablefootnote{https://civitai.com/models/16048/or-disco-elysium-style-lora}   & \textit{zaum, elysiumchar} \\
         vangogh \tablefootnote{https://civitai.com/models/91/van-gogh-diffusion}            & \textit{Starry Night by Van Gogh, lvngvncnt} \\
         \bottomrule
    \end{tabular}
    \caption{Community models used for evaluation. Each model captures a different target style.}
    \label{tab:supp_model}
\end{table}

\begin{figure}[h]
    \centering
    \includegraphics[width=1\textwidth]{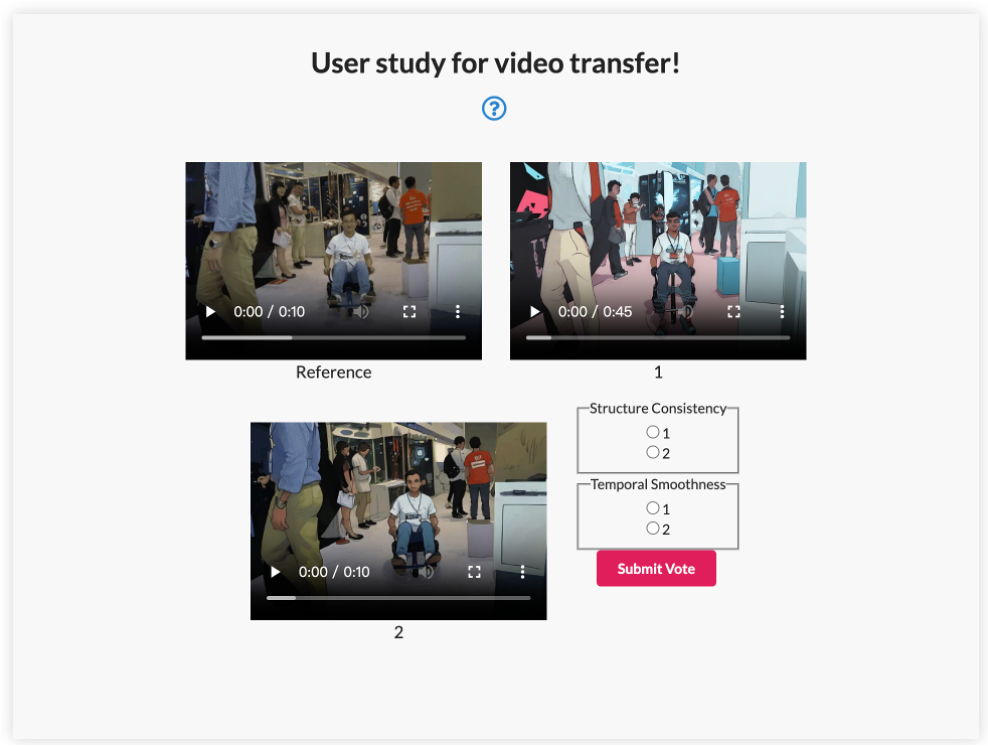}   
    \caption{In our user study, the participant is given triplets of videos: The \enquote{Reference} is the input video, that videos $1$ and $2$ should be structurally consistent with, in addition to being temporally smooth. For each of these two aspects the user chooses which of the two videos fulfills this aspect best.}
    \label{fig:supp_user_study}
\end{figure}

\subsection{Evaluation}\label{supp:eval}

\iparagraph{Models for evaluation} We use three Dreambooth and LoRA settings for evaluation, the model name and trigger words are shown in \cref{tab:supp_model}. For the evaluation, we use the trigger word as the prefix of our prompt.

\iparagraph{User study}
Our user study involved 15 participants who collectively watched a total of 93 video clips.
The video clips were the same as those used in the quantitative evaluation.
\cref{fig:supp_user_study} illustrates the user interface of our user study system:
Participants are shown the input video as reference, as well as an output from our method and an output from one random baseline method. They are asked to select the output with better temporal smoothness and structure consistency to the input. For each baseline method, we compute the win rate of our method as 
\begin{equation}
    \mathtt{ours\_win\_rate} = 1 - \frac{\mathtt{baseline\_voted}}{\mathtt{baseline\_shown}}
\end{equation}

\iparagraph{Data captioning}
We caption the DAVIS dataset with CogVLM\cite{wang2023cogvlm}, which is a state-of-the-art visual language model.
For each video clip, we feed the middle frame together with the following prompt:

\vspace{-0.1cm}
\begin{tcolorbox}[colback=white,colframe=black]
\textit{
    Please caption the given image. 
    The caption should focus on the main object in image and describe the motion of the object.
}
\end{tcolorbox}

\begin{figure}
    \begin{center}
        \includegraphics[width=1\textwidth]{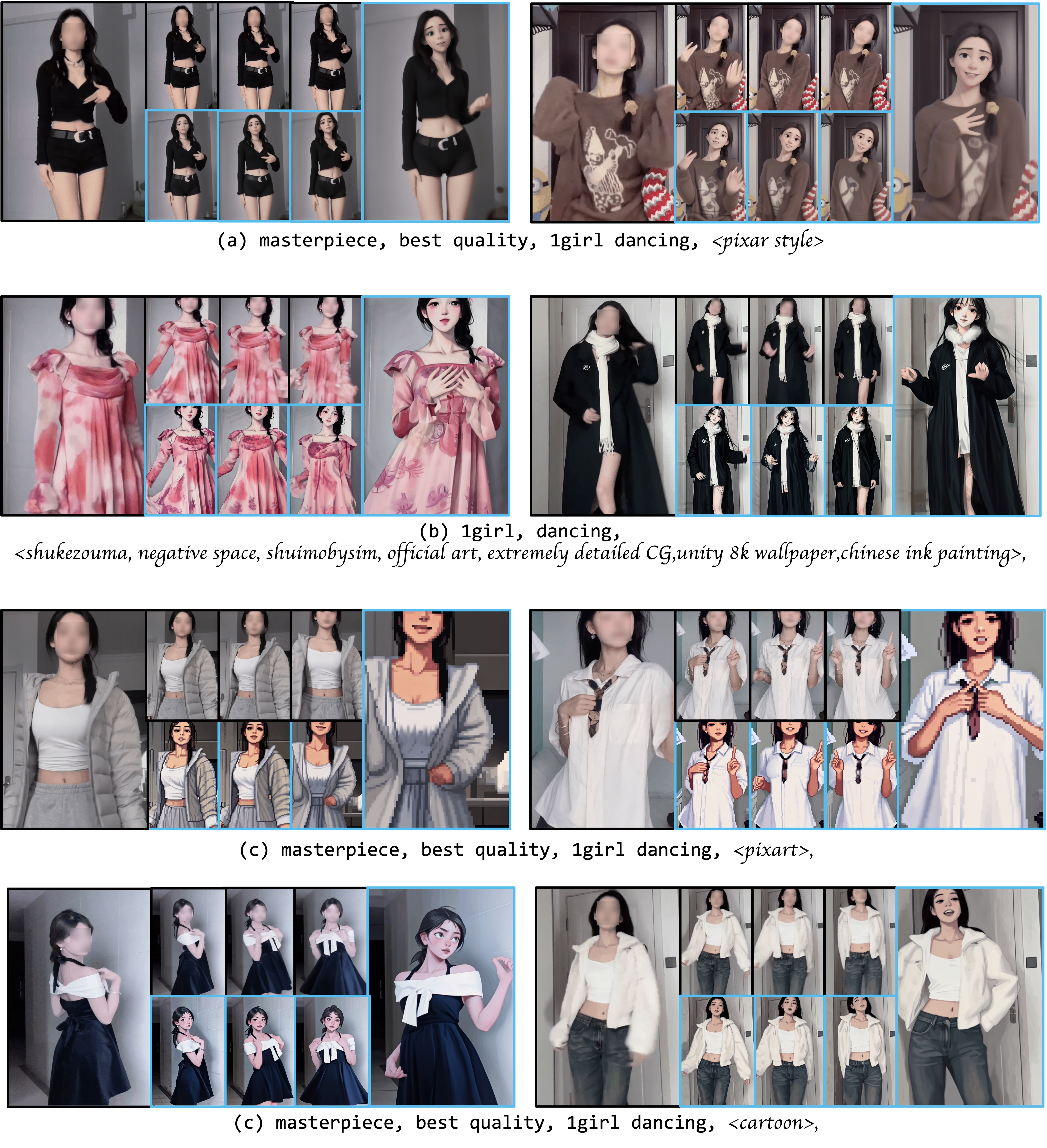}
    \end{center}
    \caption{ 
        Our method translates an input video stream (black boundaries) into an output video stream (blue boundaries) that conforms to a desired target style. Each prompt is composed of the caption obtained for the input video (see \cref{supp:eval} followed by target style.
    }
    \label{fig:app}
\end{figure}


\subsection{Application}

\cref{fig:app} shows another application of our method, demonstrating its potential in virtual-liver cases.
We transfer the input videos to different styles at 10 - 15 FPS on an NVIDIA RTX 4090.

\newpage

\end{document}